\documentclass{article}

\usepackage{PRIMEarxiv}

\usepackage[utf8]{inputenc} 
\usepackage[T1]{fontenc}    
\usepackage{hyperref}       
\usepackage{url}            
\usepackage{booktabs}       
\usepackage{amsfonts}       
\usepackage{nicefrac}       
\usepackage{microtype}      
\usepackage{lipsum}
\usepackage{graphicx}
\graphicspath{{media/}}     
\usepackage{amsmath}
\usepackage{amssymb}
\usepackage{mathtools}
\usepackage{amsthm}
\usepackage{algorithm}
\usepackage{algorithmic}
\usepackage{microtype}
\usepackage{graphicx}
\usepackage{subfigure}
\usepackage{booktabs} 

\usepackage{hyperref}

\usepackage{tablefootnote}
  
\title{The Multiple Subnetwork Hypothesis \\
\large Enabling Multidomain 
Learning by Isolating Task-Specific Subnetworks in Feedforward 
Neural Networks
}

\author{
  Jacob Renn, Ian Sotnek, Benjamin Harvey \\
  AI Squared, Inc. \\
  United States \\
  \texttt{\{jacob.renn, ian.sotnek, benjamin.harvey\}@squared.ai} \\
   \And
  Brian Caffo \\
  Department of Biostatistics \\
  Johns Hopkins University \\
  Baltimore, MD, USA\\
  \texttt{bcaffo1@jhu.edu} \\
}

\begin{document}
\maketitle

\begin{abstract}
    Neural networks have seen an explosion of usage and research in the past decade, particularly within the domains of
computer vision and natural language processing.  However, only recently have advancements in neural networks yielded performance 
improvements beyond narrow applications and translated to expanded multitask models capable of generalizing across 
multiple data types and modalities.  Simultaneously, it has been shown that neural networks are overparameterized to a high degree, 
and pruning techniques have proved capable of significantly reducing the number of 
active weights within the network while largely preserving performance.  In this work, we identify a methodology and network representational structure which allows
a pruned network to employ previously unused weights to learn subsequent tasks.  We employ these methodologies on well-known benchmarking datasets for testing 
purposes and show that networks trained using our approaches are able to learn multiple tasks, which may be related or unrelated,
in parallel or in sequence without sacrificing performance on any task or exhibiting catastrophic forgetting.
\end{abstract}


\twocolumn

\section{Introduction}
\label{Introduction}

Recent advancements in deep learning have seen artificial neural networks (ANNs) match or exceed human-level performance 
in narrow tasks \cite{single_character, mastering_go, deep_blue, starcraft}.  When exposing artificial intelligence (AI) 
to environments that are flexible and novel, however, narrow AI loses the ability to flexibly identify and react to 
patterns.  AI remains unable to match biological intelligence in flexibly negotiating dynamically changing environments.  
This is because biological learning enables the continuous acquisition of new knowledge and development of new capabilities 
across multiple disparate inputs, all without the newly learned information destroying internal state representations 
necessary to perform previously learned skills or utilize previously acquired knowledge.

The inability of a traditional ANN to learn multiple tasks across heterogeneous data sources without exhibiting catastrophic forgetting 
is a significant obstacle on the path to generalizable AI.  Solving this challenge is a step towards enabling AI to emulate 
biological continuous multidomain learning; rather than relying on separate ANNs to learn and perform multiple tasks, a single 
neural network capable of learning multiple tasks can replicate the flexibility of human learning, simplifying the deployment 
of the capabilities of an AI system in dynamic, chaotic environments \cite{self_driving,multitask_mixture}.

As demonstrated in this manuscript, the training procedure described herein outputs a single neural network containing multiple 
subnetworks, each of which is capable of learning a single task. Practically speaking, this methodology yields a single network with the capabilities 
of multiple disparate networks, while also doing so more efficiently by only utilizing a small portion of the original network's weights 
for each task.

\section{Background}

In this work, we are concerned with overcoming catastrophic forgetting in the development of ANNs capable of learning multiple 
tasks across disparate domains.  To accomplish this, we pose and investigate the \textit{Multiple Subnetwork Hypothesis} and 
build upon it to create a multitask ANN architecture capable of learning multiple tasks across various different data types and 
with different input and output shapes.

\subsection{The Multiple Subnetwork Hypothesis}

We propose the \textit{Multitple Subnetwork Hypothesis} -- that a dense, randomly-initialized feedforward ANN contains within its 
architecture multiple disjoint subnetworks which can be utilized together to learn and make accurate predictions on multiple 
tasks, regardless of the degree of similarity between tasks or input types.  Additionally, we have developed the network 
distillation procedure \textit{Reduction in Sub-Network Neuroplasticity} (RSN2), which optimizes and sparsifies dense ANNs, to 
test our hypothesis.  Our approach enables state of the art performance on multiple tasks within the same neural network without 
parameter overlap, and therefore without exposing subnetworks to the conditions giving rise to interference during multitask learning.

\begin{figure}[ht]
    \vskip 0.2in
    \begin{center}
    \centerline{\includegraphics[width=\columnwidth]{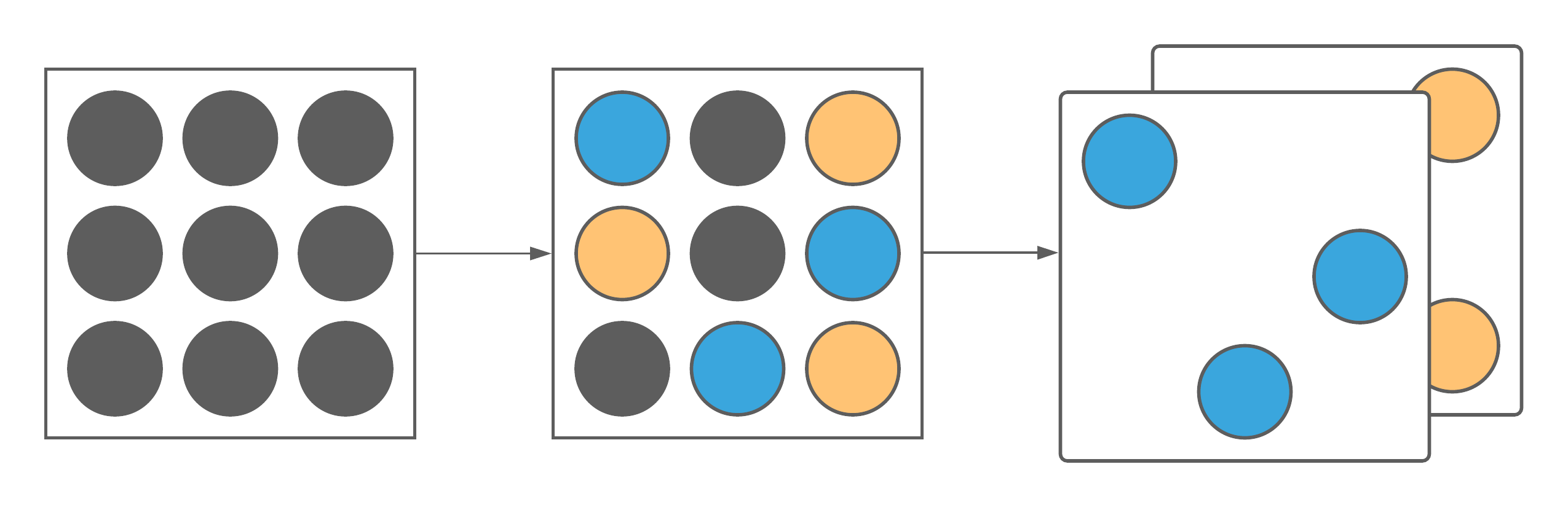}}
    \caption{Illustration of the Multiple Subnetwork Hypothesis. Multiple subnetworks able to represent a fully-connected neural network
    for a given task are identified in a single dense, randomly-initialized feedforward neural network.  These multiple specialist 
    subnetworks can then be arrayed as a sparse tensor to isolate task representations, shielding them from catastrophic 
    interference and enabling multi-domain task performance.}
    \label{fig:rsn2_introduction}
    \end{center}
    \vskip -0.2in
    \end{figure}

\subsection{Sparse Subnetwork Identification and Isolation}

It is well known and documented that ANNs are often extremely overparameterized, and this overparameterization results in redundancies and 
inefficiencies which can be eliminated to conserve computational resources without negatively affecting model accuracy \cite{optimalbraindamage}.
The pruning of an ANN to reduce or eliminate extraneous parameters yields a sparse representation of the original fully-connected network, and 
the challenge is therefore to identify which subnetwork of parameters within a dense ANN will best represent the entire original ANN \cite{suzuki2001simple,
han2015deep, lis2019full}.

One means of reducing parameter count and computational requirements without sacrificing the accuracy of the sparse network is through the identification of 
sparse subnetworks within dense neural networks. In a randomly-initialized feed-forward network, there may be a subnetwork which can sufficiently represent 
the original network, called in other works a “winning lottery ticket” \cite{lotteryticket}.  This methodology serves as the primary inspiration of our 
Multiple Subnetwork Hypothesis, as we explore in this study whether the parameters excluded from a subnetwork for one task may be utilized in the formation 
of subnetworks capable of learning and performing additional tasks. 

\subsection{Interference-Resistant Multitask Learning}

Continuous learning in both biological and artificial neural systems is constrained by the stability-plasticity dilemma. Learning must be sufficiently 
plastic to allow the acquisition of new knowledge, but also sufficiently stable to prevent the forgetting of previous knowledge \cite{mermillod2013stability}. 
Interference stemming from lack of stability during sequential learning in ANNs, often referred to as Catastrophic Forgetting, is a result of the tendency of 
NNs to destroy internal state representations used in learning previously acquired tasks when learning a new task \cite{cfinconnectionistnetworks,
goodfellow2013empirical, pfulb2019comprehensive}. This phenomenon is especially pronounced in continuous learning paradigms with a low degree of inter-task 
relatedness \cite{aljundi2017expert,ma2018modeling,masana2021ternary}, posing a challenge to creating multitask models able to perform prediction or 
classification across very different tasks, input data types, or input shapes. Effective mitigation of catastrophic forgetting is therefore a key prerequisite 
to achieving the flexibility and adaptability of a multidomain multitask artificial intelligence. 

Isolating parameters for a given task, thereby fixing parameter subsets for a given task, is a means of achieving maximal stability in a continuous learning 
context \cite{delange2021continual}. Parameter isolation methods localize the performance of learned tasks over a physically \cite{yoon2017lifelong,
rusu2016progressive,mallya2018packnet} or logically \cite{liu2018rotate,serra2018overcoming, mallya2018piggyback} isolated region of an ANN, reducing or 
eliminating the interaction of weights responsible for a given task, therefore minimizing the opportunity for the acquisition of new tasks to interfere with 
the ability of an ANN to continue to perform previously learned tasks. The limitation of interactions between functional groups of neurons has its basis in 
the biological necessity to not only functionally integrate information in the brain, but to isolate the processing of that information into distinct modules, 
as evidenced by the importance of locally specialized regions in processing sensory information \cite{zavitz2019understanding}. While the isolation of 
subnetworks is the surest way to prevent interference arising from learning new tasks, this method can scale poorly under memory and computational constraints 
and therefore demands a flexible approach to limiting network capacity \cite{yoon2017lifelong} or minimizing the size of the network utilized for each task. 
Our methodology aims to negate the high computational cost of parameter isolation by applying our RSN2 methodology to represent multidomain multitask models 
as sparse tensors, thereby minimizing their computational footprint. 

\section{Methodology}

In this section, we will discuss how we enable a single model to learn multiple tasks across multiple domains and datatypes 
through both a task-specific weight representational structure and a modified training procedure which selects disjoint subsets 
of network weights for each individual task.

\subsection{Multitask Representational Structure}

The weights of fully connected neural network layers traditionally consist of a kernel tensor, $k$, of shape $m \times n$, and a bias vector,
$b$, of length $n$, where $m$ denotes the number of columns in the input data and $n$ denotes the number of neurons within the layer.  
To process input data $\bar{x}$, the decision function is thus $\Phi(k\bar{x} + b)$, where $\Phi$ denotes the layer's activation function.
Suppose then that there are inputs from two different distributions, $\bar{x_1}$ and $\bar{x_2}$.  Given the two-dimensional structure of the 
kernel tensor in this traditional representation, there is no way to alter the decision function given the distribution currently selected.

To address this problem, we add a new dimension, $t$, to the kernel tensor and the bias vector within the layer to denote which distribution or task
the input data belongs to, leaving the kernel tensor with a shape of $t \times m \times n$ and transforming the bias vector into a bias matrix of shape
$t \times n$. The decision function for this new layer is thus altered to that in Equation \ref{eq:mt_decision_function}.

\begin{equation}
    \label{eq:mt_decision_function}
    F(\bar{x}, i) = \Phi(k_{i}\bar{x} + b_{i})
\end{equation}

We take a similar approach for convolutional layers.  For a two-dimensional convolutions with color channels, the kernel is of shape
$s_1 \times s_2 \times c \times f$, where $s_1$ and $s_2$ denote the width and height of the convolutional filters, $c$ represents the 
number of channels in the input, and $f$ corresponds to the number of filters.  The bias vector in this scenario has length $f$.  In our
multitask representation of the convolutional layer, we once more add a new dimension to the front of both the kernel tensor and the bias 
vector to denote the task.

\begin{algorithm}[tb]
    \caption{RSN2 Training Procedure}
    \label{alg:rsn2}
 \begin{algorithmic}
    \STATE {\bfseries Input:} Neural Network $F(x)$, Training data for multiple tasks, $X_{i}, Y_{i}, i \in \{1,2,..,N\}$
    \FOR{$i \in \{1,2,..,N\}$}
        \STATE{Choose $p_{i}, p_{i} \in (0, 100)$, ensuring $\sum_{i = 1}^{N}p_{i} <= 100$}
        \STATE{Deactivate all weights which have been used for previous tasks}
        \STATE{Activate all weights which have not been used for previous tasks}
        \STATE{Select $(x_{i}, y_{i})$ as a subset of training data and labels for the current task}
        \STATE{Calculate active weight gradients with respect to $(x_{i}, y_{i})$}
        \STATE{Set all inactive weight gradients to 0}
        \STATE{Identify the quantity $q_{i}$ such that $p\%$ of weight gradients with respect to $(x_{i}, y_{i})$ is less than $q_{i}$}
        \STATE{Deactivate all weights with gradients less than $q_{i}$ and do not allow them to train}
        \STATE{Train the network as usual, only altering identified active weights and keeping all other weights inactive}
        \STATE{Save all weight values for the specified task}
    \ENDFOR{}
    \STATE{At inference time for task $i$, activate only stored weights for task $i$ and perform inference as usual}
 \end{algorithmic}
 \end{algorithm}

\subsection{Sparsification and Subnetwork Identification}

While the multitask representational structure described above could theoretically be used to enable a network to learn multiple tasks 
as-is, it alone does not help us answer our hypothesis.  Alone, this structure allows multiple networks to be combined into a single model,
as the number of parameters within each layer is increased by a factor of the number of tasks. To fully test our Multiple Subnetwork Hypothesis,
we devised the RSN2 training procedure, which ensures that only one weight along the task dimension is active for all other fixed indices 
in multitask layers.

\begin{equation}
    \label{eq:indicator_function}
    I(x) = 
        \left\{\begin{array}{lr}
            1, & \text{if } x \neq 0\\
            0, & \text{otherwise}
        \end{array}\right\}
\end{equation}

Mathematically, we utilize the indicator function from Equation \ref{eq:indicator_function} as an activation function to indicate whether 
an individual weight is active or inactive within a layer.  Using this function, we can therefore demonstrate whether disjoint subnetworks 
are active for all tasks by taking the sum of the indicator function across the task dimension for all other fixed indices.  If for all 
fixed indices, $b$ and $c$, the condition present in Equation \ref{eq:multitask_indicator_condition} holds, then each weight is only active 
once across all tasks, meaning the task dimension adds no additional active weights to the network layer.

\begin{equation}
    \label{eq:multitask_indicator_condition}
    \sum_{a = 1}^{t}I(k_{a,b,c}) \leq 1; \text{ } b,c \text{ fixed}
\end{equation}

We utilize binary masking in our training procedure to impose these conditions.  These masks therefore ensure only select weights are active
during training through the bitwise multiplication of the weight value and the mask value.

\begin{figure}[ht]
    \vskip 0.2in
    \begin{center}
    \centerline{\includegraphics[width=\columnwidth]{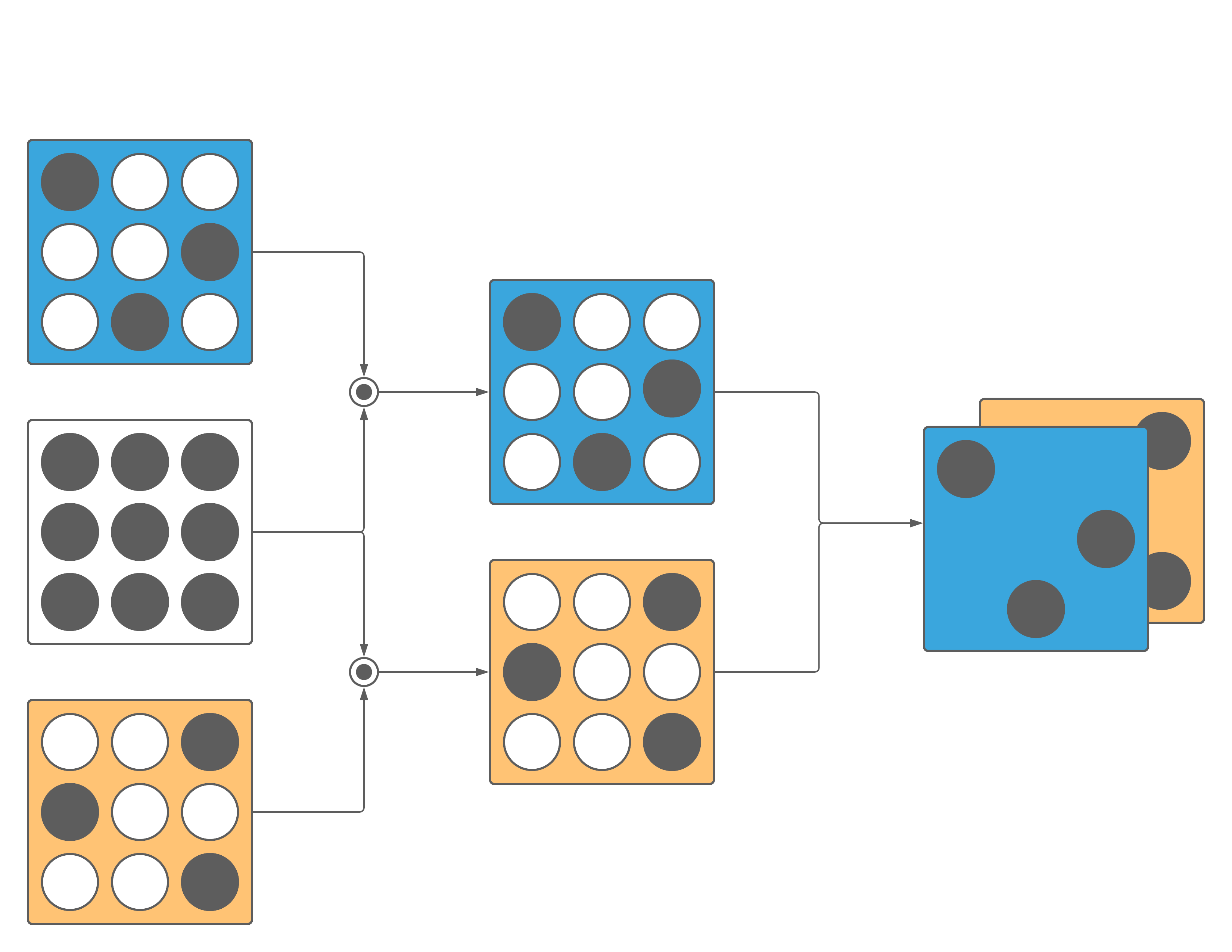}}
    \caption{Diagram of the RSN2 training procedure and multitask representational 
    structure for a single fully-connected layer with two tasks.  On the left, we begin with the initialized 
    layer weight matrix (center) and create two binary mask matrices (top and bottom).  These binary masks 
    are then bitwise-multiplied to the weight matrix, resulting in two disjoint subsets of weights which are 
    then separated across the task dimension in our representational structure.}
    \label{fig:rsn2_methods}
    \end{center}
    \vskip -0.2in
    \end{figure}

For weight selection and pruning, we utilize a gradient-based approach to identify which weights are to be selected for each task.  By 
performing pruning in this manner, all pruning is done in a one-shot manner after initialization rather than iteratively as the model 
trains, with the pruning rate a hyperparameter selected at training time.  The entirety of the RSN2 training procedure for a single 
network layer can be found in Algorithm \ref{alg:rsn2}.  For performing inference, the masks used for training can be removed, thus
reducing the number of active weights to a maximum value of $t*p*W$, where $t$ represents the number of tasks, $p$ the maximum proportion 
of weights active for any one task, and $W$ the number of weights which would be present in the network had a traditional, one-task structure 
been used.  Note that to satisfy Equation \ref{eq:mt_decision_function}, $t*p \leq 1$.  As a result, the RSN2 procedure, when coupled with 
the multitask representational structure defined above, identifies and separates disjoint subnetworks within a traditional network to allow 
each to individually perform separate tasks. Due to the lack of interaction between these subnetworks in the resulting model, similarity 
between tasks is not required, and no transfer learning will occur between tasks.  A visual representation of the RSN2 training procedure 
and multitask representational structure can be found in Figure \ref{fig:rsn2_methods}.

\subsection{Experiments}

We test our methodologies across eight experiments utilizing multiple datasets to test the Multiple Subnetwork Hypothesis and RSN2 training 
procedure.  In our first experiment, we test the merits of the hypothesis by identifying whether a single network can learn the same task 
multiple times.  To test this, we create a single convolutional multitask network and isolate five separate subnetworks within it, each 
utilizing a disjoint 20\% of the weights within the network.  Each of these subnetworks is then trained on the Fashion MNIST dataset \cite{fashion_mnist}. 
We then compare the results of each of these subnetworks, as well as the results of the ensemble consisting of all five subnetworks, against the performance 
of a dedicated network with identical architecture.

Our second experiment builds upon the first by identifying if a single network can learn related tasks from a single dataset.  The classes from the same 
Fashion MNIST dataset are broken into two groups for this experiment; the first group consists of the five individual classes which were most easily 
identified in the previous experiment.\footnote{We identified those which were learned most easily by the resulting F1 score on test data by the control model.}
The second group consists of the remaining five classes.  The results of this experiment are compared to the performance of a dedicated network with identical 
architecture.

For our third experiment, we train a fully-connected network to perform both the MNIST Digit Recognition \cite{mnist} and Fashion MNIST tasks and compare the 
performance of a single network to two dedicated networks.  This experiment is the first we perform to test whether a single network can truly learn different 
tasks from multiple different domains and distributions, and it is the first experiment designed to show whether a fully connected network can perform multiple 
tasks.

In our fourth experiment, we train a convolutional network to perform both MNIST tasks from the previous experiment.  This network is identical in architecture to 
all convolutional architectures trained previously, and once again this network is compared to two dedicated networks of the same architecture.

Our fifth, sixth, and seventh experiments continue to explore our hypothesis to a greater degree, as we train individual networks to perform multiple tasks across a 
wider array of data modalities. In the fifth experiment, we train a fully connected network to perform three tasks: both MNIST tasks as well as a regression task on 
house prices in the Boston area \cite{boston_housing}.  Experiment 6 trains a three-task convolutional network across both MNIST tasks as well as the CIFAR-10 dataset 
\cite{cifar10}.  Our seventh and final experiment tests whether a single network can learn multiple tasks across an even wider array of input data types.  For this 
experiment, we train a single fully-connected network to perform four tasks: both MNIST tasks, the Boston housing regression task, and an IMDB reviews sentiment 
classification task \cite{imdb}. 

Our final experiment seeks to explore the feasibility of our methodologies in the context of more complicated tasks. For this experiment, we trained a single convolutional 
model to perform four tasks. The first three of these are the age, gender, and race prediction tasks from the UTKFace dataset \cite{utkface}. The fourth task is the CIFAR10 
task \cite{cifar10}. This experimental network is tested against four dedicated networks of identical architectures.

\section{Results}

In this section, we present the results for each experiment conducted.  Firstly, we present the results for each control model for all tasks, as well as describe the 
model architectures and training schedules, as well as any data preprocessing steps which were taken.  For control results, each control model was created only once 
and will therefore be presented with greater detail upon first presentation, but its results will be included in every pertinent experiment.  For more detailed indicators 
of performance for all models, as well as mappings between integer labels and their respective classes, we refer the reader to the appendices.

\subsection{Experiment 1: MNIST Fashion Ensemble, Convolutional Architecture}

Our first experiment involved creating a single convolutional network with five disjoint subnetworks each trained on the MNIST Fashion dataset.  We utilized a convolutional 
architecture with two convolutional blocks, with the first block containing two convolutional layers with thirty-two $3\times3$ filters and ReLU activation \cite{relu} followed 
by maximum pooling over a $2\times2$ filter area.  The second convolutional block was identical to the first, except each of the convolutional layers utilized sixty-four filters. 
The output to the convolutional blocks were then flattened and fed into two fully-connected layers with $256$ neurons each and ReLU activation.  A final fully-connected layer with 
ten neurons and softmax activation was used to provide the final outputs.  The only preprocessing which was done on input data was a division of pixel values by $255$ to ensure all 
input values were within the interval from $0$ to $1$.  The model was trained using a batch size of $512$, and early stopping was initialized after three epochs with no improvement 
of greater than $0.01$ in loss on validation data.  The same architecture, preprocessing steps, and training procedure was completed in the experimental case as well, with each 
subnetwork pruned so that 20\% of the network weights were allocated to each task.  A summary of results can be found in Table \ref{tab:exp1}.

\begin{table}[t]
    \caption{Experiment 1 Results.}
    \label{tab:exp1}
    \vskip 0.15in
    \begin{center}
    \begin{small}
    \begin{sc}
    \begin{tabular}{lcc}
    \toprule
    Network & Accuracy & Pruning Rate \\
    \midrule
        \multicolumn{3}{c}{\textbf{Experimental}} \\ 
        \hline
        Subnetwork 1 & 90\% & 80\% \\
        Subnetwork 2 & 90\% & 80\% \\
        Subnetwork 3 & 90\% & 80\% \\
        Subnetwork 4 & 90\% & 80\% \\
        Subnetwork 5 & 92\% & 80\% \\
        Ensemble & 92\% & 80\% Each \\
        \hline
        \multicolumn{3}{c}{\textbf{Control}} \\
        \hline
        MNIST Fashion & 92\% & N/A \\
    \bottomrule
    \end{tabular}
    \end{sc}
    \end{small}
    \end{center}
    \vskip -0.1in
    \end{table}

\subsection{Experiment 2: MNIST Fashion Two-Task Model, Convolutional Architecture}

For our second experiment, we utilized an identical architecture, data preprocessing, and training procedure to the previous experiment but trained two individual subnetworks within that architecture.  We trained the 
first subnetwork on the five classes which were most easily identified by the control model, measured by F1 score, and we trained the second subnetwork on the five classes 
which were the most difficult for the control model to identify, also measured by F1 score.  The results of the experiment can be found in Table \ref{tab:exp2}.

\begin{table}[t]
    \caption{Experiment 2 results.}
    \label{tab:exp2}
    \vskip 0.15in
    \begin{center}
    \begin{small}
    \begin{sc}
    \begin{tabular}{lcc}
    \toprule
    Network & Accuracy & Pruning Rate \\
    \midrule
        \multicolumn{3}{c}{\textbf{Experimental}} \\
        \hline
        Easy Task & 98\% & 80\% \\
        Hard Task & 84\% & 80\% \\
        \hline
        \multicolumn{3}{c}{\textbf{Control}} \\
        \hline
        MNIST Fashion & 92\% & N/A \\
    \bottomrule
    \end{tabular}
    \end{sc}
    \end{small}
    \end{center}
    \vskip -0.1in
    \end{table}

\subsection{Experiment 3: MNIST Digits and MNIST Fashion Model, Fully-Connected Architecture}

Our third experiment is the first experiment to truly test whether a single model can learn disparate tasks across a variety of data modalities.  To do this, we trained a single fully-connected network to 
perform both the MNIST Digit Recognition and the MNIST Fashion Recognition tasks.  For this experiment, we preprocessed the data for both tasks once more by dividing each pixel value by $255$.  Each image 
was then flattened into a two-dimensional vector.  The images would then be passed through six fully connected layers each containing $1000$ artificial neurons each and activated using the ReLU activation 
function.  The final layer contained ten neurons which were activated with the softmax activation function.  The model was trained with the same early stopping conditions as all previous models, and a batch 
size of $512$ was used once more.  The experimental model was trained using the same procedure and with the same architecture, with each task pruned to utilize 10\% of the network weights.  The results of 
this experiment can be found in Table \ref{tab:exp3}.

\begin{table}[t]
    \caption{Experiment 3 results.}
    \label{tab:exp3}
    \vskip 0.15in
    \begin{center}
    \begin{small}
    \begin{sc}
    \begin{tabular}{lcc}
    \toprule
    Network & Accuracy & Pruning Rate \\
    \midrule
        \multicolumn{3}{c}{\textbf{Experimental}} \\
        \hline
        MNIST Digit & 97\% & 90\% \\
        MNIST Fashion & 86\% & 90\% \\
        \hline
        \multicolumn{3}{c}{\textbf{Control}} \\
        \hline  
        MNIST Digit & 97\% & N/A \\
        MNIST Fashion & 88\% & N/A \\
    \bottomrule
    \end{tabular}
    \end{sc}
    \end{small}
    \end{center}
    \vskip -0.1in
    \end{table}

\subsection{Experiment 4: MNIST Digit and MNIST Fashion Model, Convolutional Architecture}

In experiment four, we repeated the modeling goals of the previous experiment, but this experiment utilized the convolutional architecture from the previous experiments, once again with an 80\% pruning rate 
for each task.  Results for this experiment can be found in Table \ref{tab:exp4}.

\begin{table}[t]
    \caption{Experiment 4 results.}
    \label{tab:exp4}
    \vskip 0.15in
    \begin{center}
    \begin{small}
    \begin{sc}
    \begin{tabular}{lcc}
    \toprule
    Network & Accuracy & Pruning Rate \\
    \midrule
        \multicolumn{3}{c}{\textbf{Experimental}} \\
        \hline
        MNIST Digit & 99\% & 80\% \\
        MNIST Fashion & 90\% & 80\% \\
        \hline
        \multicolumn{3}{c}{\textbf{Control}} \\
        \hline
        MNIST Digit & 99\% & N/A \\
        MNIST Fashion & 92\% & N/A \\
    \bottomrule
    \end{tabular}
    \end{sc}
    \end{small}
    \end{center}
    \vskip -0.1in
    \end{table}

\subsection{Experiment 5: MNIST Digit, MNIST Fashion, and Boston Housing Model, Fully-Connected Architecture}

The fifth experiment provides the greatest test for our Multiple Subnetwork Hypothesis up to this point, as we train a model with a wider variety of datatypes and tasks, and we also train some tasks in sequence rather than all 
at once.  For this experiment, we trained a single fully-connected network first on both MNIST tasks, then trained the same network on the Boston Housing dataset.  The MNIST portions of the network were identical in architecture 
to the ones used in previous experiments, but due to difference in input shapes between the MNIST images and the Boston Housing dataset\footnote{The MNIST images are both 28 pixels wide and 28 pixels tall, while the Boston Housing 
dataset contains only 13 input features.} an additional fully-connected layer is added after the Boston Housing input layer to create shape alignment.  This layer is activated with ReLU activation, and it is also pruned in 
the experimental model.  In the test case, all subnetworks are pruned to only use 10\% of available weights.

Data preprocessing for MNIST tasks is identical to previous MNIST preprocessing for fully-connected networks.  For the Boston Housing dataset, input features were scaled such that the minimum feature value in the training data 
corresponded to a value of $0$ and the maximum value of the feature in the training data corresponded to a value of $1$.  The predicted values were also scaled using the same procedures. 
During training, as stated before, both MNIST tasks are learned simultaneously, utilizing the same early stopping criteria and batch sizes as the previous cases.  For the Boston Housing task, the same early stopping procedure is 
used, but the batch size is decreased to $32$ in both the control and test cases due to the limited amount of data available.  The results for this experiment can be found in Table \ref{tab:exp5}.

\begin{table}[t]
    \caption{Experiment 5 results.  In the performance column, percentages correspond to accuracy percentage on test data, while decimals represent mean squared error on test data.}
    \label{tab:exp5}
    \vskip 0.15in
    \begin{center}
    \begin{small}
    \begin{sc}
    \begin{tabular}{lcc}
    \toprule
    Network & Performance & Pruning Rate \\
    \midrule
        \multicolumn{3}{c}{\textbf{Experimental}} \\
        \hline
        MNIST Digit & 97\% & 90\% \\
        MNIST Fashion & 87\% & 90\% \\
        Boston Housing & 0.016 & 90\% \\
        \hline
        \multicolumn{3}{c}{\textbf{Control}} \\
        \hline
        MNIST Digit & 97\% & N/A \\
        MNIST Fashion & 88\% & N/A \\
        Boston Housing & 0.017 & N/A \\
    \bottomrule
    \end{tabular}
    \end{sc}
    \end{small}
    \end{center}
    \vskip -0.1in
    \end{table}

\begin{figure}[ht]
    \vskip 0.2in
    \begin{center}
    \centerline{\includegraphics[width=\columnwidth]{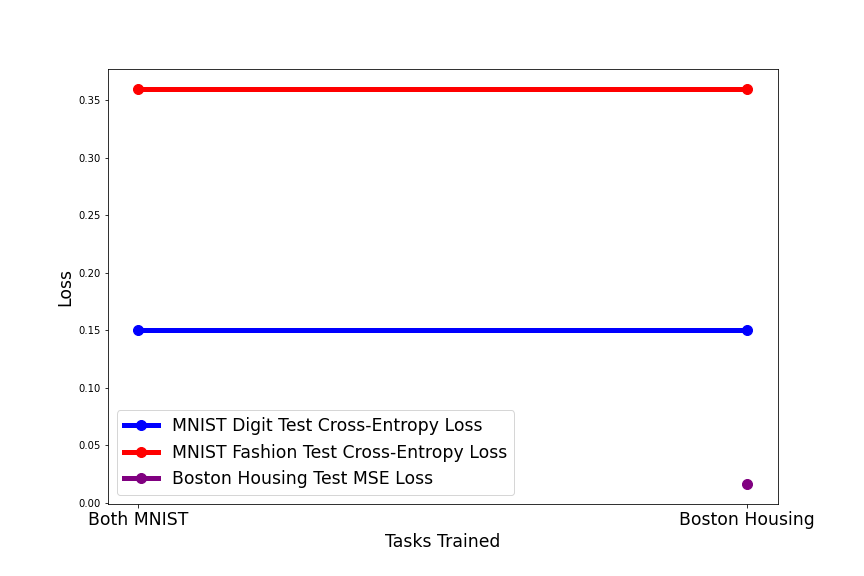}}
    \caption{Task losses obtained on test data for multitask model after training each set of tasks.}
    \label{fig:exp5}
    \end{center}
    \vskip -0.2in
\end{figure}

Furthermore, to identify whether any forgetting occurred after sequential training, we recorded loss values on test data for current and previously-trained tasks after each subsequent task was trained. 
These results are visualized in Figure \ref{fig:exp5}, where it can be seen that no performance degradation occurred on the first two tasks after training the third task.

\subsection{Experiment 6: CIFAR10, MNIST Digit, and MNIST Fashion Model, Convolutional Architecture}

\begin{figure}[ht]
    \vskip 0.2in
    \begin{center}
    \centerline{\includegraphics[width=\columnwidth]{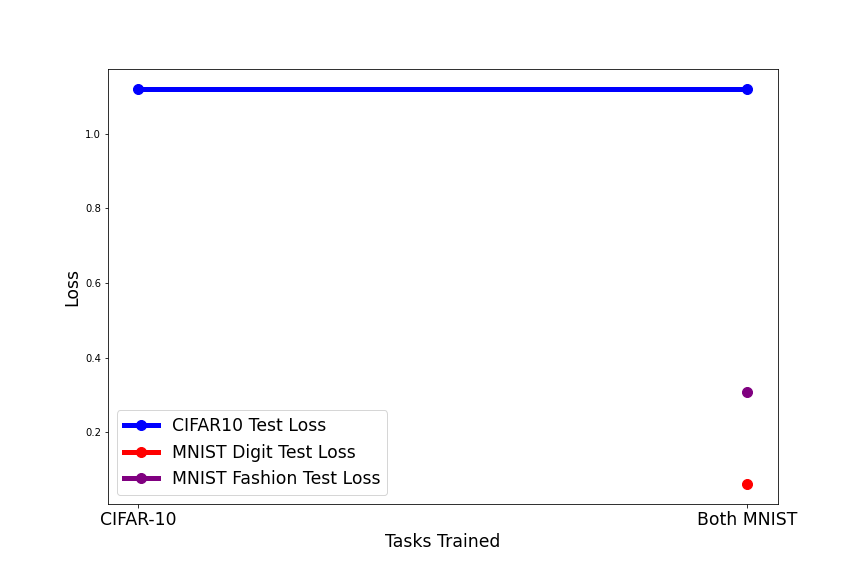}}
    \caption{Task losses obtained on test data for multitask model after training each set of tasks.}
    \label{fig:exp6}
    \end{center}
    \vskip -0.2in
\end{figure}

Experiment 6 was designed to expand upon the findings of previous experiments by testing whether a single convolutional network could converge across multiple classification tasks across a variety of 
input images.  In this experiment, we test whether a single convolutional network can converge on first the CIFAR-10 image classification task and then train the same network on both MNIST tasks we have 
utilized to this point.  To ensure that all images are of the same shape, we downsize the CIFAR-10 images so that they have the same height and width as the MNIST data.  Additionally, because of the color 
channels present in the CIFAR images, a separate first convolutional layer must be created to process the CIFAR-10 inputs in the multitask model.  This dedicated layer is still pruned to the same degree as 
all other processing layers within the model.  During training, the same early stopping procedures that have been used previously were employed, and batch sizes of $512$ were used for all tasks.  The results 
for this experiment can be found in Table \ref{tab:exp6}, and task losses after training can be found in Figure \ref{fig:exp6}.  The test model was pruned to 90\% in this experiment.

\begin{table}[t]
    \caption{Experiment 6 results.}
    \label{tab:exp6}
    \vskip 0.15in
    \begin{center}
    \begin{small}
    \begin{sc}
    \begin{tabular}{lcc}
    \toprule
    Network & Accuracy & Pruning Rate \\
    \midrule
        \multicolumn{3}{c}{\textbf{Experimental}} \\
        \hline
        CIFAR-10 & 60\% & 80\% \\
        MNIST Digit & 98\% & 90\% \\
        MNIST Fashion & 89\% & 90\% \\
        \hline
        \multicolumn{3}{c}{\textbf{Control}} \\
        \hline
        CIFAR-10 & 47\% & N/A \\
        MNIST Digit & 99\% & N/A \\
        MNIST Fashion & 92\% & N/A \\
    \bottomrule
    \end{tabular}
    \end{sc}
    \end{small}
    \end{center}
    \vskip -0.1in
\end{table}

\subsection{Experiment 7: MNIST Digit, MNIST Fashion, Boston Housing, and IMDB Reviews Model, Fully-Connected Architecture}

\begin{figure}[ht]
    \vskip 0.2in
    \begin{center}
    \centerline{\includegraphics[width=\columnwidth]{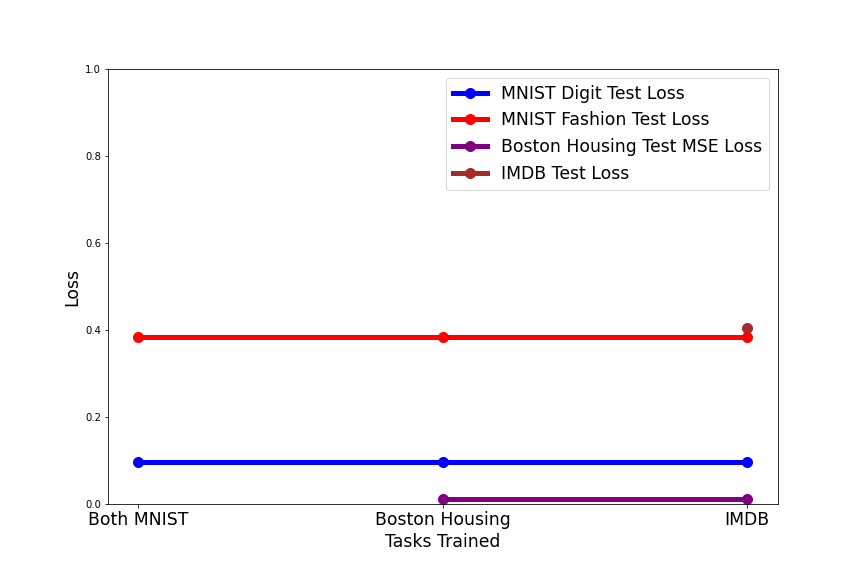}}
    \caption{Task losses obtained on test data for multitask model after training each set of tasks.}
    \label{fig:exp7}
    \end{center}
    \vskip -0.2in
\end{figure}

Our seventh experiment provides perhaps the most wide-ranging set of tasks for a single network, and as a result 
it is perhaps the best test of our Multiple Subnetwork Hypothesis.  In this experiment, we utilized a single 
fully-connected neural network and trained the network on four separate tasks; the first two of these tasks, 
both the MNIST Digit and MNIST Fashion tasks, were trained simultaneously.  The third task, the Boston Housing 
regression task, was then trained.  Finally, the IMDB reviews sentiment classification task was trained.  
The same architecture as previously used in all fully-connected experiments was used in this experiment.  For 
the MNIST and Boston Housing tasks, the same preprocessing and reshaping efforts took place as previous experiments 
with fully-connected architectures.  For the IMDB task, only the most common $10000$ words were utilized and 
sequences were padded (or truncated) to $128$ words, with both padding and truncating occurring at the end of the 
review.  Furthermore, each of the input tokens was passed through an embedding layer, which embedded the token in a 
two-dimensional vector space.  The embedded tensors were then flattened into a two-dimensional vector and passed through 
the architecture used throughout this experiment\footnote{This architecture involves six fully-connected layers of $1000$ 
neurons and ReLU activation.}.  For the multitask model in this experiment, the first fully-connected layer which processed 
inputs from the IMDB task was dedicated solely to the IMDB task due to the differences in input shapes between all tasks.  
During training, the same early stopping criteria were used for all training iterations, batches sizes of $512$ were used 
for the MNIST and IMDB tasks, while a batch size of $32$ was used for the Boston Housing task.  The results for this experiment 
can be found in Table \ref{tab:exp7}, and task losses can be found in Figure \ref{fig:exp7}.  The model was pruned such that 
each task utilizes 10\% of the available weights in fully-connected layers; the embedding layer was not pruned.

\begin{table}[t]
    \caption{Experiment 7 results. In the performance column, percentages correspond to accuracy percentage on test data, while decimals represent mean squared error on test data.}
    \label{tab:exp7}
    \vskip 0.15in
    \begin{center}
    \begin{small}
    \begin{sc}
    \begin{tabular}{lcc}
    \toprule
    Network & Performance & Pruning Rate \\
    \midrule
        \multicolumn{3}{c}{\textbf{Experimental}} \\
        \hline
        MINST Digit & 97\% & 90\% \\
        MNIST Fashion & 87\% & 90\% \\
        Boston Housing & 0.011 & 90\% \\
        IMDB & 82\% & 90\% \\
        \hline
        \multicolumn{3}{c}{\textbf{Control}} \\
        \hline
        MNIST Digit & 97\% & N/A \\
        MNIST Fashion & 88\% & N/A \\
        Boston Housing & 0.017 & N/A \\
        IMDB & 80\% & N/A \\
    \bottomrule
    \end{tabular}
    \end{sc}
    \end{small}
    \end{center}
    \vskip -0.1in
\end{table}

\subsection{Experiment 8: UTKFace and CIFAR10 Four-Task Model, Convolutional Architecture}

In our final experiment, we tested the RSN2 training procedure on a larger network with more complex tasks involved. We created a single 
network with a convolutional architecture on four individual tasks, three tasks from the UTKFace dataset \cite{utkface} and one multiclass 
classification task from CIFAR10 \cite{cifar10}.  For the age task in the UTKFace data, we grouped values into classes by decade, excluding 
a final class containing all individuals with age greater than 90.

Architecturally, the model trained in this experiment contains three convolutional blocks containing  of $16$, $32$, and $64$ $3\times3$ filters, 
respectively.  Each block contains two convolutional layers with padding to preserve shape, followed by a max pooling layer.  These blocks then 
feed into a fully connected architecture of three layers, each with $128$ neurons each and activated using ReLU activation \cite{relu}, followed 
by an output layer of the required shape to perform the specified task.

For the multitask model, we utilized a shared convolutional embedding between all UTKFace tasks. This resulted in a two-task structure for 
convolutional layers within the model, with one task channel processing the UTKFace images and the other task channel processing the CIFAR10 images.  
For the fully-connected layers, each of the UTKFace tasks was isolated and a four-task architecture was used.  Along both the convolutional and 
the fully-connected architectures, each task was pruned to utilize only $25\%$ of the model's available weights per task, meaning $50\%$ of the weights in the 
convolutional layers were utilized\footnote{This is due to the two-task structure of the convolutional layers, as mentioned above.} and $100\%$ of the weights 
across the shared fully-connected architecture was utilized.\footnote{This is due to the four-task structure of the fully-connected layers, as mentioned above.}  
All four tasks were trained simultaneously, with early stopping criteria utilized with patience of five epochs. A summary of the results of this 
experiment can be found in \ref{tab:exp8}. In addition to these performance results, which interestingly showed that the dedicated control 
models were unable to converge but that the multitask model was able to converge, it was also found that the multitask model only utilized 
$130MB$ of disk space when saved, while the combined single-task models utilized $203MB$ of disk space, over $56\%$ more than the 
multitask model.

\begin{table}[t]
    \caption{Experiment 8 results.}
    \label{tab:exp8}
    \vskip 0.15in
    \begin{center}
    \begin{small}
    \begin{sc}
    \begin{tabular}{lcc}
    \toprule
    Network & Accuracy & Pruning Rate \\
    \midrule
        \multicolumn{3}{c}{\textbf{Experimental}} \\
        \hline
        UTKFace Age & 55\% & 75\% \\
        UTKFace Gender & 89\% & 75\% \\
        UTKFace Race & 77\% & 75\% \\
        CIFAR10 & 50\% & 75\% \\
        \hline
        \multicolumn{3}{c}{\textbf{Control}} \\
        \hline
        UTKFace Age & 32\% & N/A \\
        UTKFace Gender & 48\% & N/A \\
        UTKFace Race & 14\% & N/A \\
        CIFAR10 & 10\% & N/A \\
    \bottomrule
    \end{tabular}
    \end{sc}
    \end{small}
    \end{center}
    \vskip -0.1in
\end{table}

\section{Discussion}

This study was designed to provide insight into answering our Multiple Subnetwork Hypothesis, which 
states that a dense, randomly-initialized feedforward neural network contains within its architecture 
multiple disjoint subnetworks which can be isolated to each separately learn and make accurate predictions 
on multiple tasks, regardless of the degree of similarity between tasks or input types.  To test this 
hypothesis, we developed the RSN2 procedure as well as a customized neural network representational structure 
designed to isolate individual subnetworks within the overall architecture of the global model on a 
layer-by-layer basis.  We then utilized these techniques in a series of seven experiments across a range of 
datasets utilizing both fully-connected and convolutional architectures.

Across each of these experiments, we consistently saw that a single network was able to converge on multiple 
tasks by utilizing only a small fraction of network weights for each task.  Furthermore, multitask models were 
able to perform each task with near identical performance to dedicated models with all weights active, even 
exceeding dedicated model performance in a few cases. Additionally, we saw that there needed to be no 
similarity between tasks, as in our most extreme case we were able to teach a single model to perform two 
separate image classification tasks, a tabular regression task, and a sentiment classification task on natural 
language.

The beneficial implications of this approach arise from the simplicity of interacting with a single model rather 
than a collection of models as in the traditional machine learning deployment paradigm. Take for example the 
human visual system: a unified stream of visual information is collected and passed to the optic lobe, where it 
is then passed to the relevant regions of the brain for further downstream analysis. This paradigm is reflected 
in the unified data pipeline of a multitask model trained using the RSN2 procedure, where invoking the model 
involves invoking the relevant task or tasks automatically. Therefore, a collection of models capable of 
performing multiple tasks can be deployed as a single entity within an organization's IT infrastructure, 
thereby improving efficiency and lowering costs. Take, for instance, a simple example of an organization 
which wants to deploy five separate models trained on five separate tasks. A typical deployment scenario would 
involve this organization deploying one model each to five separate computational nodes. Using a model trained 
with the RSN2 procedure to perform all five tasks, this organization could therefore reduce costs by deploying 
this single multitask model to a single node, saving 80\% of the organization's original costs. Furthermore, 
to improve system robustness by introducing redundancy, the organization could deploy this single model 
to a set of three nodes and still reduce costs by 40\%.

Additionally, in our final experiment, even though we do not utilize sparse tensors to save the model, the 
pruned multitask model utilizes less disk space when saved compared to the total amount of the identical 
single-task models. This indicates that utilizing RSN2 and our multitask representational structure already 
leads to more efficient representation of the models themselves, thus potentially further reducing costs by 
decreasing the hardware requirements needed to save and use models for inference.

\subsection{Limitations}

There are a number of limitations to this study which limit the conclusions we can draw from the results.  
Firstly, our experiments focus primarily on computer vision related tasks in this work, with all experiments 
containing some image classification task.  While we do also include experiments which test our methods 
on both tabular input and natural language inputs, further studies should still be conducted to more rigorously 
test our hypotheses and methods on various input and data types.

Another limitation to this study is the limited size of both the datasets and models which were studied.  Many 
state of the art models have been trained on millions of data samples, and the models themselves consist of 
many tens of millions of parameters.  Our study, on the other hand, considered labeled data with up to tens of 
thousands of samples and models which contained hundreds of thousands to millions of parameters, orders of magnitude 
fewer (in both cases) than the state of the art in most cases.  It is therefore unknown whether these methods would 
need to be modified\footnote{In particular, it is unknown whether our one-shot sparsification technique would 
generalize to much larger models.} to accommodate larger modeling paradigms. 

\subsection{Conclusions}

Despite the limitations to this study, there are still a number of conclusions we can draw from its results.  
Each multitask model was clearly able to converge on all its assigned tasks, achieving performance similar to, 
and in some cases exceeding, the control cases each time.  These results therefore confirm our Multiple Subnetwork 
Hypothesis, showing that a single neural architecture is capable of utilizing small portions of itself to learn 
individual tasks.  Additionally, we showed that our methodology for identifying and logically separating network 
weights is robust to catastrophic forgetting, and we showed that we were therefore able to train multitask models 
across various domains and tasks both in parallel or in sequence without any adverse effects on previously-learned 
tasks.

Because the resulting model would have the ability to robustly perform multiple tasks, we can also conclude that 
a single model trained with these techniques could substitute multiple models in a deployed scenario.  This 
substitution would therefore reduce resource requirements for organizations deploying and utilizing models, thus 
saving costs and simplifying the deployment architectures these organizations use.  These models would also be able 
to acquire more downstream tasks after previously being trained on initial tasks, meaning adding new predictive 
analytics would be a much simpler undertaking as well, as it would only involve updating an already-deployed model 
and not require creating a completely new deployment for every new analytic.

\subsection{Future Work}

Following studies should be designed to address some of the limitations of this work as well as to 
improve the methods used within.  Future studies should provide a more thorough analysis on how 
these methods perform on a larger variety of data modalities, including data types, task types, data 
availability, and model sizes.  Secondly, we intend to identify whether a more precise pruning method 
can be applied as training occurs instead of in a one-shot manner as was done in this study.  If pruning 
can be applied in a more intelligent manner, we believe the resulting model would potentially be able to 
more consistently surpass unpruned model performance while also achieving greater sparsification rates than 
possible with one-shot techniques.

Additionally, this study does not perform any analysis to identify the maximum amount of tasks which can be 
learned by a single network.  Future work should therefore address finding the upper bounds of the number of tasks 
an individual model can support.  Finally, the methods in this work do not allow for any transfer learning across 
tasks; instead, it is assumed that new tasks should utilize a completely disjoint subset of weights relative to all 
previously-learned tasks.  Additional studies should be conducted to identify whether transfer learning can be 
leveraged during the acquisition of similar tasks to enable further reduction in the number of active weights and 
the quicker acquisition of new tasks.


\newpage
\onecolumn
\bibliographystyle{unsrt}  
\bibliography{example_paper}  

\newpage
\appendix
\onecolumn
\section{Class labels for classification tasks.}
\label{app:labels}

\begin{table}[th!]
    \caption{Class labels for computer vision tasks.}
    \label{tab:cv_classes}
    \vskip 0.15in
    \begin{center}
    \begin{small}
    \begin{sc}
    \begin{tabular}{lcc}
    \toprule
    Dataset & Integer & Label \\
    \midrule
     \hline
     MNIST Digit & & \\
     \hline
     & 1 & Digit ``0'' \\
     & 2 & Digit ``1'' \\
     & 3 & Digit ``2'' \\
     & 4 & Digit ``3'' \\
     & 5 & Digit ``4'' \\
     & 6 & Digit ``5'' \\
     & 7 & Digit ``6'' \\
     & 8 & Digit ``7'' \\
     & 9 & Digit ``8'' \\
     & 10 & Digit ``9'' \\
     \hline
     MNIST Fashion & & \\
     \hline
     & 1 & T-shirt/top \\
     & 2 & Trouser \\
     & 3 & Pullover \\
     & 4 & Dress \\
     & 5 & Coat \\
     & 6 & Sandal \\
     & 7 & Shirt \\
     & 8 & Sneaker \\
     & 9 & Bag \\
     & 10 & Ankle Boot \\
     \hline
     CIFAR-10 & & \\
     \hline
     & 1 & Airplane \\
     & 2 & Automobile \\
     & 3 & Bird \\
     & 4 & Cat \\
     & 5 & Deer \\
     & 6 & Dog \\
     & 7 & Frog \\
     & 8 & Horse \\
     & 9 & Ship \\
     & 10 & Truck \\
     \hline
     UTKFace Age & & \\
     \hline
     & 1 & $0-10$ \\
     & 2 & $11-20$ \\
     & 3 & $21-30$ \\
     & 4 & $31-40$ \\
     & 5 & $41-50$ \\
     & 6 & $51-60$ \\
     & 7 & $61-70$ \\
     & 8 & $71-80$ \\
     & 9 & $81-90$ \\
     & 10 & $91+$ \\
     \hline
     UTKFace Gender & & \\
     & 1 & Male \\
     & 2 & Female \\
     \hline
     UTKFace Ethnicity & & \\
     & 1 & White \\
     & 2 & Black \\
     & 3 & Asian \\
     & 4 & Indian \\
     & 5 & ``Other'' \tablefootnote{Defined by dataset creators and includes ethnicities such as Hispanic, Latino, and Middle Eastern.} \\
    \bottomrule
    \end{tabular}
    \end{sc}
    \end{small}
    \end{center}
    \vskip -0.1in 
\end{table}

\begin{table}[th]
    \caption{Class labels for IMDB task.}
    \label{tab:imdb_classes}
    \vskip 0.15in
    \begin{center}
    \begin{small}
    \begin{sc}
    \begin{tabular}{cc}
    \toprule
    Integer & Label \\
    \midrule
        1 & Negative Review \\
        2 & Positive Review \\
    \bottomrule
    \end{tabular}
    \end{sc}
    \end{small}
    \end{center}
    \vskip -0.1in 
\end{table}

\newpage
\onecolumn
\section{Additional model performance metrics.}
\label{app:metrics}

When conducting our study, we collected additional performance measures for all models which were not 
included in the main manuscript above.  These metrics are presented in the following tables.

\begin{table}[th]
    \caption{Per-class F1 and overall accuracy scores for control computer vision models. ``FC'' indicates fully-connected architecture. ``Conv'' indicates convolutional architecture.}
    \label{tab:control_vision}
    \vskip 0.15in
    \begin{center}
    \begin{small}
    \begin{sc}
    \begin{tabular}{lccccccccccc}
    \toprule
    \multicolumn{1}{c}{Network} & \multicolumn{10}{c}{Class Number} & Accuracy \\
    \midrule
     & 1 & 2 & 3 & 4 & 5 & 6 & 7 & 8 & 9 & 10 & \\
     \hline
     MNIST Digit FC & 0.98 & 0.99 & 0.97 & 0.97 & 0.98 & 0.97 & 0.98 & 0.97 & 0.96 & 0.96 & 0.97 \\
     MNIST Digit Conv & 0.99 & 0.99 & 0.98 & 0.99 & 0.99 & 0.99 & 0.99 & 0.99 & 0.99 & 0.98 & 0.99 \\
     MNIST Fashion FC & 0.84 & 0.98 & 0.79 & 0.88 & 0.79 & 0.97 & 0.69 & 0.95 & 0.97 & 0.96 & 0.88 \\
     MNIST Fashion Conv & 0.85 & 0.99 & 0.87 & 0.92 & 0.86 & 0.98 & 0.76 & 0.97 & 0.99 & 0.97 & 0.92 \\
     CIFAR-10 Conv (Exp 6) & 0.55 & 0.59 & 0.35 & 0.26 & 0.38 & 0.39 & 0.51 & 0.53 & 0.55 & 0.52 & 0.47 \\
     UTKFace Age Conv & 0.00 & 0.00 & 0.49 & 0.00 & 0.00 & 0.00 & 0.00 & 0.00 & 0.00 & 0.00 & 0.32 \\
     UTKFace Gender Conv & 0.00 & 0.64 & N/A & N/A & N/A & N/A & N/A & N/A & N/A & N/A & 0.48 \\
     UTKFace Ethnicity Conv & 0.00 & 0.00 & 0.25 & 0.00 & 0.00 & N/A & N/A & N/A & N/A & N/A & 0.14 \\
     CIFAR-10 Conv (Exp 8) & 0.00 & 0.00 & 0.00 & 0.00 & 0.00 & 0.18 & 0.00 & 0.00 & 0.00 & 0.00 & 0.10 \\
    \bottomrule
    \end{tabular}
    \end{sc}
    \end{small}
    \end{center}
    \vskip -0.1in
\end{table}

\begin{table}[th]
    \caption{Per-class F1 and overall accuracy scores for experimental computer vision models.}
    \label{tab:exp_vision}
    \vskip 0.15in
    \begin{center}
    \begin{small}
    \begin{sc}
    \begin{tabular}{lccccccccccc}
    \toprule
    \multicolumn{1}{c}{Network} & \multicolumn{10}{c}{Class Number} & Accuracy \\
    \midrule
     & 1 & 2 & 3 & 4 & 5 & 6 & 7 & 8 & 9 & 10 & \\
     \hline
     \textbf{Experiment 1} \\
     \hline
     Subnetwork 1 & 0.86 & 0.98 & 0.85 & 0.90 & 0.84 & 0.98 & 0.72 & 0.96 & 0.97 & 0.97 & 0.90 \\
     Subnetwork 2 & 0.85 & 0.97 & 0.84 & 0.89 & 0.83 & 0.98 & 0.72 & 0.96 & 0.98 & 0.96 & 0.90 \\
     Subnetwork 3 & 0.84 & 0.98 & 0.82 & 0.88 & 0.83 & 0.98 & 0.72 & 0.96 & 0.97 & 0.97 & 0.90 \\
     Subnetwork 4 & 0.85 & 0.98 & 0.84 & 0.90 & 0.84 & 0.98 & 0.72 & 0.96 & 0.98 & 0.97 & 0.90 \\
     Subnetwork 5 & 0.87 & 0.99 & 0.88 & 0.91 & 0.88 & 0.98 & 0.78 & 0.96 & 0.98 & 0.97 & 0.92 \\
     Ensemble & 0.87 & 0.98 & 0.87 & 0.91 & 0.86 & 0.98 & 0.76 & 0.97 & 0.98 & 0.97 & 0.92 \\
     \hline
     \textbf{Experiment 2} \\
     \hline
     Easier Task Subnetwork & N/A & 1.00 & N/A & N/A & N/A & 0.97 & N/A & 0.96 & 0.99 & 0.96 & 0.98 \\
     Harder Task Subnetwork & 0.85 & N/A & 0.87 & 0.92 & 0.85 & N/A & 0.73 & N/A & N/A & N/A & 0.84 \\
     \hline
     \textbf{Experiment 3} \\
     \hline
     MNIST Digit & 0.99 & 0.99 & 0.98 & 0.97 & 0.97 & 0.97 & 0.98 & 0.98 & 0.97 & 0.96 & 0.98 \\
     MNIST Fashion & 0.82 & 0.97 & 0.77 & 0.87 & 0.79 & 0.96 & 0.62 & 0.94 & 0.97 & 0.95 & 0.87 \\
     \hline
     \textbf{Experiment 4} \\
     \hline
     MNIST Digit & 0.99 & 0.99 & 0.99 & 0.99 & 0.99 & 0.99 & 0.99 & 0.99 & 0.98 & 0.98 & 0.99 \\
     MNIST Fashion & 0.85 & 0.98 & 0.85 & 0.90 & 0.85 & 0.98 & 0.71 & 0.96 & 0.97 & 0.96 & 0.90 \\ 
     \hline
     \textbf{Experiment 5} \\
     \hline
     MNIST Digit & 0.98 & 0.99 & 0.97 & 0.97 & 0.97 & 0.97 & 0.97 & 0.98 & 0.97 & 0.96 & 0.97 \\
     MNIST Fashion & 0.82 & 0.98 & 0.78 & 0.88 & 0.78 & 0.95 & 0.68 & 0.94 & 0.96 & 0.95 & 0.87 \\
     \hline
     \textbf{Experiment 6} \\
     \hline
     MNIST Digit & 0.98 & 0.99 & 0.98 & 0.98 & 0.98 & 0.98 & 0.98 & 0.98 & 0.98 & 0.97 & 0.98 \\ 
     MNIST Fashion & 0.84 & 0.98 & 0.83 & 0.89 & 0.83 & 0.97 & 0.70 & 0.95 & 0.97 & 0.96 & 0.89 \\
     CIFAR-10 & 0.65 & 0.72 & 0.46 & 0.43 & 0.49 & 0.49 & 0.69 & 0.70 & 0.70 & 0.69 & 0.60 \\
     \hline
     \textbf{Experiment 7} \\
     \hline
     MNIST Digit & 0.98 & 0.99 & 0.98 & 0.96 & 0.97 & 0.96 & 0.97 & 0.96 & 0.96 & 0.96 & 0.97 \\ 
     MNIST Fashion & 0.82 & 0.97 & 0.78 & 0.87 & 0.78 & 0.95 & 0.64 & 0.93 & 0.95 & 0.95 & 0.87 \\
     \hline
     \textbf{Experiment 8} \\
     \hline
     UTKFace Age & 0.89 & 0.45 & 0.68 & 0.37 & 0.14 & 0.44 & 0.34 & 0.29 & 0.45 & 0.08 & 0.55 \\
     UTKFace Gender & 0.90 & 0.88 & N/A & N/A & N/A & N/A & N/A & N/A & N/A & N/A & 0.89 \\
     UTKFace Ethnicity & 0.84 & 0.83 & 0.79 & 0.68 & 0.27 & N/A & N/A & N/A & N/A & 0.77 \\
     CIFAR-10 & 0.58 & 0.63 & 0.27 & 0.35 & 0.44 & 0.47 & 0.53 & 0.54 & 0.61 & 0.52 & 0.50 \\
    \bottomrule
    \end{tabular}
    \end{sc}
    \end{small}
    \end{center}
    \vskip -0.1in
\end{table}

\begin{table}[th]
    \caption{Per-class F1 and overall accuracy scores for experimental and control IMDB sentiment analysis models.}
    \label{tab:imdb}
    \vskip 0.15in
    \begin{center}
    \begin{small}
    \begin{sc}
    \begin{tabular}{lccc}
    \toprule
    \multicolumn{1}{c}{Network} & \multicolumn{2}{c}{Class Number} & Accuracy \\
    \midrule
        & 1 & 2 & \\
        \hline
        Experiment 7 IMDB & 0.83 & 0.81 & 0.82 \\
        Control IMDB & 0.80 & 0.80 & 0.80 \\
    \bottomrule
    \end{tabular}
    \end{sc}
    \end{small}
    \end{center}
    \vskip -0.1in
\end{table}

\begin{table}[th]
    \caption{Boston Housing model losses.}
    \label{tab:boston}
    \vskip 0.15in
    \begin{center}
    \begin{small}
    \begin{sc}
    \begin{tabular}{lc}
    \toprule
    Network & Loss \\
    \midrule
        Experiment 5 & 0.016 \\
        Experiment 7 & 0.011 \\
        Control & 0.017 \\
    \bottomrule
    \end{tabular}
    \end{sc}
    \end{small}
    \end{center}
    \vskip -0.1in
\end{table}

\end{document}